\documentclass{article}

\usepackage[preprint]{neurips_2019}

\usepackage[utf8]{inputenc} 
\usepackage[T1]{fontenc}    
\usepackage{hyperref}       
\usepackage{url}            
\usepackage{booktabs}       
\usepackage{amsfonts}       
\usepackage{nicefrac}       
\usepackage{microtype}      

\usepackage{multicol}
\usepackage{comment}
\usepackage{wrapfig}

\usepackage{soul}
\usepackage[dvipsnames]{xcolor}
\usepackage{listings}
\usepackage{marginnote} 
\usepackage{mathtools, amsthm, amsmath, amssymb}
\usepackage{natbib}
\usepackage{standalone}
\usepackage{tikz}
\usetikzlibrary{shapes,arrows}
\usetikzlibrary{bayesnet}
\usetikzlibrary{arrows, decorations.markings}
\usepackage{graphicx}
\usepackage{subcaption}

\usepackage{enumitem}

\definecolor{darkred}{RGB}{208, 25, 8}
\definecolor{dkgreen}{rgb}{0,0.4,0}
\definecolor{dkblue}{rgb}{0,0.1,0.5}

\renewcommand{\cite}{\citep}

\usepackage{ragged2e}

\definecolor{keywordblue}{RGB}{25,25,117}
\definecolor{stringred}{RGB}{159, 34, 34}
\definecolor{commentgreen}{RGB}{29, 138, 29}
\definecolor{emphpurple}{RGB}{79,12,58}

\newcommand{\listingsfont}{\footnotesize\ttfamily}
\usepackage{algorithm}
\usepackage[noend]{algpseudocode} 

\makeatletter
\newcommand\fs@nobottomruled{\def\@fs@cfont{\bfseries}\let\@fs@capt\floatc@ruled
	\def\@fs@pre{\hrule height.8pt depth0pt \kern2pt}%
	\def\@fs@post{}
	\def\@fs@mid{\kern2pt\hrule\kern2pt}%
	\let\@fs@iftopcapt\iftrue}
\makeatother

\floatstyle{nobottomruled}
\restylefloat{algorithm}

\renewcommand{\line}{\noindent\makebox[\linewidth]{\rule{\columnwidth}{0.1pt}}}

\algnewcommand{\IfThen}[2]{
	\State \algorithmicif\ #1\ \algorithmicthen\ #2\ }
\algnewcommand{\Arguments}{\item[\textbf{Arguments:}]}
\algnewcommand{\Returns}{\item[\textbf{Returns:}]}

\newcommand{\alg}[1]{\textsc{#1}}

\newcommand{\vparams}{\boldsymbol{\theta}}
\newcommand{\pparams}{\boldsymbol{\lambda}}

\newcommand{\loss}{\mathcal{L}}
\DeclareMathOperator*{\argmax}{arg\,\!max}

\newcommand{\kw}[1]{\mbox{\lstinline$#1$}}
\newcommand{\kwsmall}[1]{\mbox{\lstset{basicstyle=\ttfamily\linespread{0.8}\scriptsize}\lstinline$#1$}}

\newcommand{\params}{\mathbf{z}}

\newcommand{\data}{\mathbf{x}}
\newcommand{\normal}{\mathcal{N}}

\newcommand{\ess}{\mathrm{ESS}/\nabla}

\usepackage{hyperref}
\hypersetup{
	colorlinks = true,
	linkcolor = darkred,
	anchorcolor = dkblue,
	citecolor = darkred,
	filecolor = dkblue,
	menucolor = dkgreen,
	runcolor = dkblue,
	urlcolor = dkblue,
}

\newcommand{\m}[1]{\mathrm{\scalebox{0.9}{#1}}}

\title{Automatic Reparameterisation of Probabilistic Programs}

\author{Maria I. Gorinova \thanks{Work done while interning at Google.} \\
	University of Edinburgh\\
	Edinburgh, UK \\
	\texttt{m.gorinova@ed.ac.uk} \\
	\And
	Dave Moore \\
	Google \\
	San Francisco, CA \\
	\texttt{davmre@google.com} \\
	\And
	Matthew D. Hoffman \\
	Google \\
	San Francisco, CA \\
	\texttt{mhoffman@google.com} \\
}

\begin{document}
\maketitle
\begin{abstract}
Probabilistic programming has emerged as a powerful paradigm in statistics, applied science, and machine learning: by decoupling modelling from inference, it promises to allow modellers to directly reason about the processes generating data. However, the performance of inference algorithms can be dramatically affected by the parameterisation used to express a model, requiring users to transform their programs in non-intuitive ways. We argue for automating these transformations, and demonstrate that mechanisms available in recent modeling frameworks can implement non-centring and related reparameterisations. This enables new inference algorithms, and we propose two: a simple approach using interleaved sampling and a novel variational formulation that searches over a continuous space of parameterisations. We show that these approaches enable robust inference across a range of models, and can yield more efficient samplers than the best fixed parameterisation.
\end{abstract}

\section{Introduction}

Reparameterising a probabilistic model means expressing it in terms of new variables defined by a bijective transformation of the original variables of interest. The reparameterised model expresses the same statistical assumptions as the original, but can have drastically different posterior geometry, with significant implications for both variational and sampling-based inference algorithms.

\begin{figure*}
	\centering
	\begin{subfigure}[b]{0.35\textwidth}
		\centering
		\includegraphics[width=\textwidth]{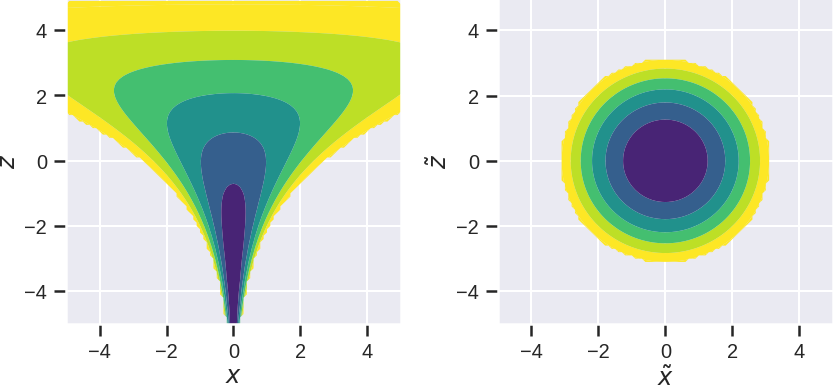}
		\caption{Centred (left) and non-centred (right) parameterisation.}
		\label{fig:funnel_intro}
	\end{subfigure}
	\hspace{1em}%
	\begin{subfigure}[b]{0.23\textwidth} 
		\begin{align*}
		\mathrm{Ne}&\mathrm{alsFunnel}(z,~x): \\
		& z \sim \normal(0, 3) \\
		& x \sim \normal(0, \exp(z / 2)) \\
		\end{align*}
		\caption{Model that generates variables $z$ and $x$.}
		\label{fig:model}
	\end{subfigure}%
	\hspace{1em}%
	\begin{subfigure}[b]{0.3\textwidth} 
		\begin{align*}
		& z = 0 \\
		& \mathrm{lp}_z = \log p_{\normal}(z \mid 0, 3) \\
		& x = 0 \\ 
		& \mathrm{lp}_x = \log p_{\normal}(x \mid 0, \exp(z / 2))
		\end{align*}
		\caption{The model in the context of \lstinline{log_prob_at_0}.}
		\label{fig:model_handled}
	\end{subfigure}
	\caption{Neal's funnel \citep{NealsFunnel}: $z \sim N(0, 3);$ $x \sim N(0, e^{z/2})$.}
	\vspace{-12pt}
\end{figure*}

Non-centring is a particularly common form of reparameterisation in Bayesian hierarchical models. Consider a random variable $z \sim \normal(\mu, \sigma)$; we say this is in  \textit{centred} parameterisation (CP). If we instead work with an auxiliary, standard normal variable $\tilde{z} \sim \normal(0,1)$, and obtain $z$ by applying the transformation $z = \mu + \sigma \tilde{z}$, we say the variable $\tilde z$ is in its \textit{non-centred} parameterisation (NCP).
Although the centred parameterisation is often more intuitive, non-centring can dramatically improve the performance of inference \citep{BetancourtCPvsNCP}. \textit{Neal's funnel} (\autoref{fig:funnel_intro}) provides a simple example: most Markov chain Monte Carlo (MCMC) algorithms have trouble sampling from the funnel due to the strong non-linear dependence between latent variables. Non-centring the model removes this dependence, converting the funnel into a spherical Gaussian distribution.

Bayesian practitioners are often advised to manually non-centre their models \cite{stan2016stan}; however, this breaks the separation between modelling and inference and requires expressing the model in a potentially less intuitive form. Moreover, it requires the user to understand the concept of non-centring and to know {\em a priori} where in the model it might be appropriate. Because the best parameterisation for a given model may vary across datasets, 
even experts may need to find the optimal parameterisation by trial and error, burdening modellers and slowing the model development loop \cite{blei2014build}.
 
We propose that non-centring and similar reparameterisations be handled \textit{automatically} by probabilistic programming systems. We demonstrate how such program transformations may be implemented using the effect handling mechanisms present in several modern deep probabilistic programming frameworks, and consider two inference algorithms enabled by automatic reparameterisation: interleaved Hamiltonian Monte Carlo (iHMC), which alternates HMC steps between centred and non-centred parameterisations, and a novel algorithm we call Variationally Inferred Parameterisation (VIP), which searches over a continuous space of reparameterisations that includes non-centring as a special case.\footnotemark~
Experiments demonstrate that these strategies enable robust inference, performing at least as well as the best fixed parameterisation across a range of models, and sometimes better, without requiring {\em a priori} knowledge of the optimal parameterisation.

\footnotetext{Code for these algorithms and experiments is available at \url{https://github.com/mgorinova/autoreparam}
}

\section{Related work}

The value of non-centring is well-known to MCMC practitioners and researchers \citep{stan2016stan, BetancourtCPvsNCP}, and can also lead to better variational fits in hierarchical models \cite{EvalVI}. However, the literature largely treats this as a modelling choice; \citet{EvalVI} propose that ``there is no general rule to determine whether non-centred parameterisation is better than the centred one.'' We are not aware of prior work that treats non-centring directly as a computational phenomenon to be exploited by inference systems.

Non-centred parameterisation of probabilistic models can be seen as analogous to the reparameterisation trick in stochastic optimisation \citep{kingma2013auto}; both involve expressing a variable in terms of a diffeomorphic transformation from a "standardised" variable. In the context of probabilistic inference, these are complementary tools: the reparameterisation trick yields low-variance stochastic gradients of variational objectives, whereas non-centring changes the geometry of the posterior itself, leading to qualitatively different variational fits and MCMC trajectories.

In the context of Gibbs sampling, \citet{papaspiliopoulos2007general} introduce a family of \textit{partially non-centred} parameterisations equivalent to those we use in VIP (described below) and show that it improves mixing in a spatial GLMM. Our current work can be viewed as an extension of this work that mechanically reparameterises user-provided models and automates the choice of parameterisation. Similarly, \citet{ASIS} proposed a Gibbs sampling scheme that interleaves steps in centered and non-centered parameterisations; our interleaved HMC algorithm can be viewed as an automated, gradient-based descendent of their scheme.

Recently, there has been work on accelerating MCMC inference through learned reparameterisation: \citet{Parno2018Transport} and \citet{hoffman2019neutra} run samplers in the image of a bijective map fitted to transform the target distribution approximately to an isotropic Gaussian. These may be viewed as `black-box’ methods that rely on learning the target geometry, potentially using highly expressive neural variational models, while we use probabilistic-program transformations to apply `white-box’ reparameterisations similar to those a modeller could in principle implement themselves. Because they exploit model structure, white-box approaches can correct pathologies such as those of Neal's funnel (\autoref{fig:funnel_intro}) directly, reliably, and at much lower cost (in parameters and inference overhead) than black-box models. White- and black-box reparameterisations are not mutually exclusive, and may have complementary advantages; combining them is a likely fruitful direction for improving inference in structured models.

\section{Understanding the effects of parameterisation}

Non-centring reparameterisation is not always optimal; its usefulness depends on properties of both the model and the observed data. In this section, we work with a simple hierarchical model for which we can derive the posterior analytically. Consider a simple realisation of a model discussed by \citet[(2)]{BetancourtCPvsNCP}, where for a vector of $N$ datapoints $\mathbf{y}$, and some given constants $\sigma$ and $\sigma_{\mu}$, we have: 
\[\theta \sim \normal(0, 1) \qquad \mu \sim \normal(\theta, \sigma_{\mu}) \qquad y_n \sim \normal(\mu, \sigma) \text{ for all } n \in 1 \dots N\]

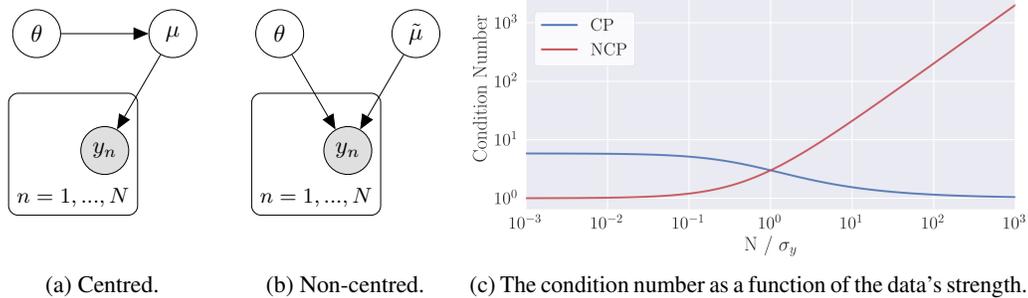
\begin{figure*}\centering
\begin{subfigure}[b]{0.225\textwidth}
	\centering
	\begin{tikzpicture}[scale=1,transform shape,
state/.style={circle,draw,thick,loop above,inner sep=0,minimum width=10}]
    \node[latent, xshift = -1cm] (theta) {$\theta$} ; %
    \node[latent, xshift = 1cm] (mu) {$\mu$} ; 
    \node[obs, below=of mu, xshift = -1cm] (y) {$y_n$} ; %
    \coordinate [yshift = -1cm] (hide) {} ;
    \edge {theta} {mu} ; %
    \edge {mu} {y} ; %
    \plate {} { %
    	(y)(hide)
    } {$n = 1, ..., N$} ;
\end{tikzpicture}
	
	\vspace{16pt}
	\caption{Centred.}
	\label{fig:cp_model}
\end{subfigure}
\begin{subfigure}[b]{0.225\textwidth}
	\centering
	\begin{tikzpicture}[scale=1,transform shape,
state/.style={circle,draw,thick,loop above,inner sep=0,minimum width=10}]
    \node[latent, xshift = -1cm] (theta) {$\theta$} ; %
    \node[latent, xshift = 1cm] (mu) {$\tilde{\mu}$} ; 
    \node[obs, below=of mu, xshift = -1cm] (y) {$y_n$} ; %
    \coordinate [yshift = -1cm] (hide) {} ;
    \edge {theta} {y} ; %
    \edge {mu} {y} ; %
    \plate {} { %
    	(y)(hide)
    } {$n = 1, ..., N$} ;
\end{tikzpicture}
	
	\vspace{16pt}
	\caption{Non-centred.}
	\label{fig:ncp_model}
\end{subfigure}
\begin{subfigure}[b]{0.53\textwidth}
	\centering
	\includegraphics[width=\textwidth]{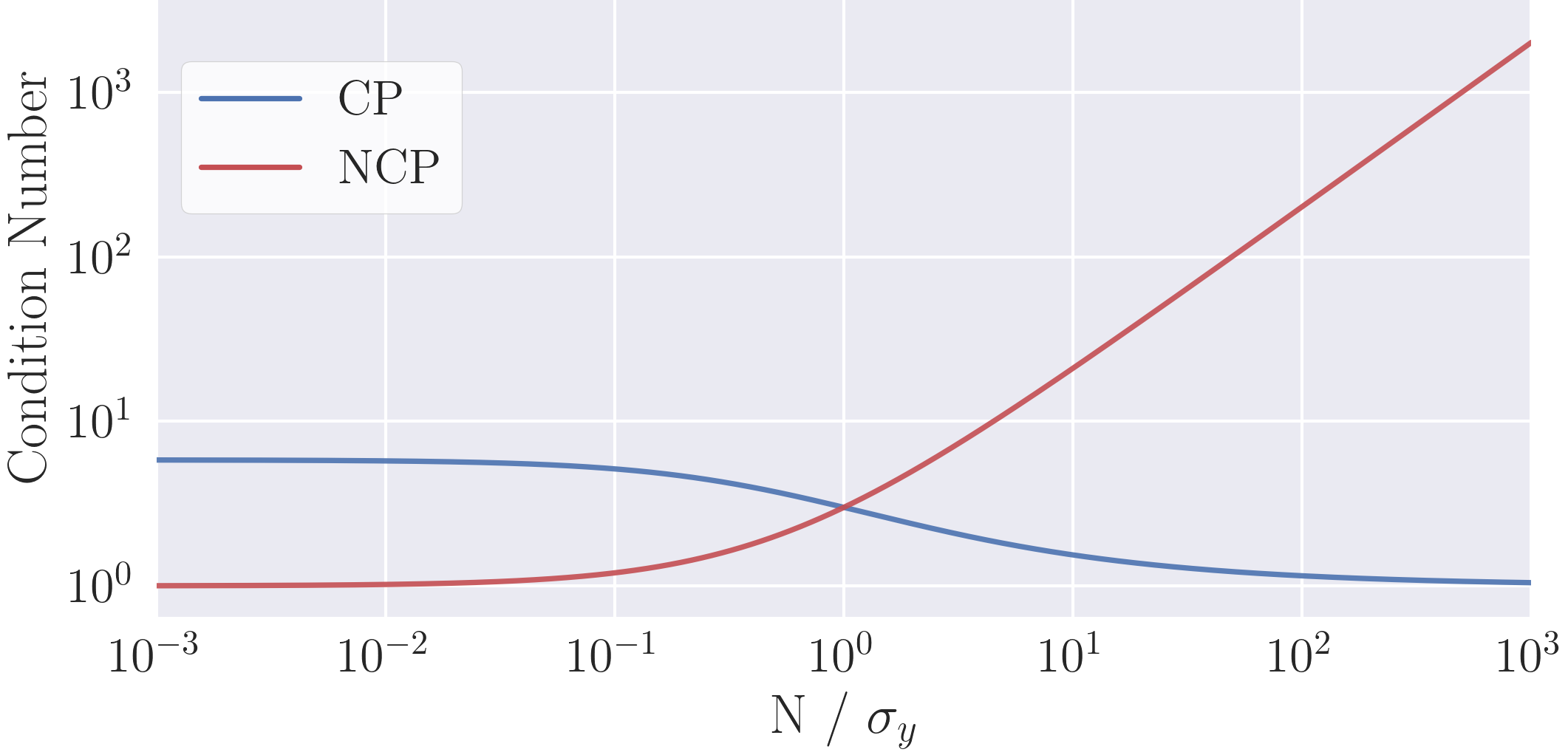}
	\caption{The condition number as a function of the data's strength.}
	\label{fig:cond_numbers}
\end{subfigure}

\caption{Effects of reparameterising a simple model with known posterior.}
\end{figure*}

In the non-centred model, $\mathbf{y}$ is defined in terms of $\tilde{\mu}$ and $\theta$, where $\tilde{\mu}$ is a standard Gaussian variable: \[\theta \sim \normal(0, 1) \qquad \tilde{\mu} \sim \normal(0, 1) \qquad y_n \sim \normal(\theta + \sigma_{\mu}\tilde{\mu}, \sigma) \text{ for all } n \in 1 \dots N\]

\autoref{fig:cp_model} and \autoref{fig:ncp_model} show the graphical models for the two parameterisations. In the non-centred case, the direct dependency between $\theta$ and $\mu$ is substituted by a conditional dependency given the data $\mathbf{y}$, which creates an ``explaining away'' effect. Intuitively, this means that the stronger the evidence $\mathbf{y}$ is (large $N$, and small variance), the stronger the dependency between $\theta$ and $\tilde{\mu}$ becomes, creating a poorly-conditioned posterior that may slow inference.

As the Gaussian distribution is self-conjugate, the posterior distribution in each case (centred or non-centred) is also a Gaussian distribution, and we can analytically inspect its covariance matrix $V$. To quantify the quality of the parameterisation in each case, we investigate the condition number $\kappa$ of the posterior covariance matrix under the optimal diagonal preconditioner. This models the common practice (implemented as the default in tools such as PyMC3 and Stan and followed in our experiments) of sampling using a fitted diagonal preconditioner.

\autoref{fig:cond_numbers} shows the condition numbers $\kappa_{\alg{cp}}$ and $\kappa_{\alg{ncp}}$ for each parameterisation as a function of $q = N / \sigma^2$; the full derivation is in \autoref{ap:maths}. This figure confirms the intuition that the non-centred parameterisation is better suited for situation when the evidence is weak, while strong evidence calls for centred parameterisation.
In this example we can exactly determine the optimal parameterisation, since the model has only one variable that can be reparameterised and the posterior has a closed form.
In more realistic settings, even experts cannot predict the optimal parameterisation for hierarchical models with many variables and groups of data, and the wrong choice can lead not just to poor conditioning but to heavy tails or other pathological geometry.

\section{Reparameterising probabilistic programs}

An advantage of probabilistic programming is that the program itself provides a structured model representation, and we can explore model reparameterisation through the lens of program transformations.  In this paper, we focus on transforming \textit{generative probabilistic programs} where the program represents a sampling process describing how the data was generated from some unknown latent variables. Most probabilistic programming languages (PPLs) provide some mechanism for transforming a generative process into an inference program; our automatic reparameterisation approach is applicable to PPLs that transform generative programs using \textit{effect handling}. This includes modern deep PPLs such as Pyro \cite{Pyro} and Edward2 \cite{Edward2}.

\subsection{Effect handling-based probabilistic programming} \label{ssec:intro_eppls}

Consider a generative program, where running the program forward generates samples from the prior over latent variables and data. Effect handling-based PPLs treat generating a random variable within such a model as an \textit{effectful operation} (an operation that is understood as having side effects) and provide ways for resolving this operation in the form of \textit{effect handlers}, to allow for inference.
For example, we often need to transform a statement that generates a random variable to a statement that evaluates some (log) density or mass function. We can implement this using an effect handler:
\[\kw{log_prob_at_0} = \kw{handler}~\{v \sim \mathcal{D}(a_1, \dots, a_N) \mapsto v = 0; \mathrm{lp}_v = \log p_{\mathcal{D}}(v \mid a_1, \dots, a_N) \}\footnotemark\]
The handler $\kw{log_prob_at_0}$ handles statements of the form $v \sim \mathcal{D}(a_1, \dots, a_N)$. The meaning of such statements is normally ``sample a random variable from the distribution $\mathcal{D}(a_1,\dots,a_N)$ and record its value in $v$''.
However, when executed in the context of $\kw{log_prob_at_0}$ (we write \lstinline|with log_prob_at_0 handle model|), statements that contain random-variable constructions are \textit{handled} by setting the value of the variable $v$ to $0$, then evaluating the log density (or mass) function of $\mathcal{D}(a_1, \dots, a_N)$ at $v=0$ and recording its value in a new (program) variable $\mathrm{lp}_v$.

\footnotetext{Typically, algebraic effects and handlers involve passing a continuation explicitly within the handler. Here, we make the continuation implicit to stay close to Edward2's implementation.}

For example, consider the function implementing Neal's funnel in \autoref{fig:model}. When executed without any context, this function generates two random variables, $z$ and $x$. When executed in the context of the $\kw{log_prob_at_0}$ handler, it does not generate random variables, but it instead evaluates $\log p_{\normal}(z \mid 0, 3)$ and $\log p_{\normal}(x \mid 0, \exp(z/2))$ (\autoref{fig:model_handled}).

This approach can be extended to produce a function that corresponds to the log joint density (or mass) function of the latent variables of the model. In \autoref{ap:make_log_joint}, we give the pseudo-code implementation of a function \lstinline|make_log_joint|, which takes a model $M(\params \mid \data)$
--- that generates latent variables $\params$ and generates and observes data $\data$ ---
and returns the function $f(\params) = \log p(\params, \data)$.
This is a core operation, as it transforms a generative model into a function proportional to the posterior distribution, which can be  repeatedly evaluated and automatically differentiated to perform inference.

More generally, effectful operations are operations that can have side effects, e.g. writing to a file. The programming languages literature formalises cases where impure behaviour arises from a set of effectful operations in terms of algebraic effects and their handlers \cite{AlgebraicEffects2001, Plotkin2009, Pretnar2015}. A concrete implementation for an effectful operation is given in the form of effect handlers, which (similarly to exception handlers) are responsible for resolving the operation. 
Effect handlers can be used as a powerful abstraction in probabilistic programming, as discussed previously by \citet{ProbProg18}, and shown by both Pyro and Edward2. 

\subsection{Model reparameterisation using effect handlers} \label{ssec:autoreparam_eppls}

Once equipped with an effect handling-based PPL, we can easily construct handlers to perform many model transformations, including model reparameterisation.

\paragraph{Non-centring handler.} $\kw{ncp} = \kw{handler}~\{ v \sim \normal(\mu, \sigma), v \notin \mathrm{data} \mapsto \tilde{v} \sim \normal(0, 1); v = \mu + \sigma\tilde{v} \}$

A non-centring handler can be used to non-centre all standardisable
\footnote{We focus on Gaussian variables, but non-centring is broadly applicable, e.g.~to the location-scale family and random variables that can be expressed as a bijective transformation $z = f_\theta(\tilde{z})$ of a ``standardised'' variable $\tilde{z}$.}
latent variables in a model. 
The handler simply applies to statements of the form $v \sim \normal(\mu, \sigma)$, where $v$ is not a data variable, and transforms them to $\tilde{v} \sim \normal(0, 1)$, $v = \mu + \sigma\tilde{v}$. When nested within a \lstinline|log_prob| handler (like the one from \autoref{ssec:intro_eppls}),  \lstinline|log_prob| handles the transformed standard normal statement $\tilde{v} \sim \normal(0, 1)$. Thus, \lstinline|make_log_joint| applied to a model in the \lstinline|ncp| context returns the log joint function of the transformed variables $\tilde{\params}$ rather than the original variables $\params$.

For example, $\kw{make_log_joint}(\m{NealsFunnel}(z,y))$ corresponds to
$\log p(z, x) = \log \normal(z \mid 0, 3) + \log \normal(y \mid 0, \exp(z / 2))$. But, 
$\kw{make_log_joint}(\kw{with ncp handle}\,\m{NealsFunnel}(z,y))$ gives the function 
$\log p(\tilde{z}, \tilde{x}) = \log \normal(\tilde{z} \mid 0, 1) + \log \normal(\tilde{x} \mid 0, 1)$, 
where $z = 3\tilde{z}$ and $x = \exp(z / 2)\tilde{x}$.

This approach can easily be extended to other parameterisations, including partially centred parameterisations (as shown later in \autoref{ssec:vip}), non-centring and whitening multivariate Gaussians, and transforming constrained variables to have unbounded support.

\paragraph{Edward2 implementation.}

We implement reparameterisation handlers in Edward2, a deep PPL embedded in Python and TensorFlow \citep{Edward2}. A model in Edward2 is a Python function that generates random variables. 
%
In the core of Edward2 is a special case of effect handling called \textit{interception}. To obtain the joint density of a model, the language provides the function \lstinline|make_log_joint_fn(model)|, which uses a \lstinline|log_prob| interceptor (handler) as previously described. 

We extend the usage of interception to treat sample statements in one parameterisation as sample statements in another parameterisation (similarly to the $\kw{ncp}$ handler above):
\begin{lstlisting}
def noncentring_interceptor(rv_constructor, **rv_kwargs):
	# Assumes rv_constructor is in the location-scale family
	name = rv_kwargs["name"] + "_std"
	rv_std = ed.interceptable$\footnotemark$(rv_constructor)(loc=0, scale=1)
	return rv_kwargs["loc"] + rv_kwargs["scale"] * rv_std
\end{lstlisting}
\footnotetext{Wrapping the constructor with \lstinline{ed.interceptable} ensures that we can nest this interceptor in the context of other interceptors.}

We use the interceptor by executing a model of interest within the interceptor's context (using Python's context managers). This overrides each random variable's constructor to construct a variable with location $0$ and scale $1$, and scale and shift that variable appropriately:
\begin{lstlisting}
with ed.interception(noncentring_interceptor): neals_funnel()
\end{lstlisting}
We present and explain in more detail all interceptors used for this work in  \autoref{ap:interceptors}. 

\section{Automatic model reparameterisation}

We introduce two inference strategies that exploit automatic reparameterisation: interleaved Hamiltonian Monte Carlo (iHMC), and the Variationally Inferred Parameterisation (VIP). 

\subsection{Interleaved Hamiltonian Monte Carlo}

Automatic reparameterisation opens up the possibility of algorithms that exploit multiple parameterisations of a single model. We consider \textit{interleaved Hamiltonian Monte Carlo (iHMC)}, which uses two HMC steps to produce each sample from the target distribution: the first step is made in CP, using the original model latent variables, while the second step is made in NCP, using the auxiliary standardised variables. Interleaving MCMC kernels across parameterisations has been explored in previous work on Gibbs sampling \cite{ASIS, VolatilityASIS}, which demonstrated that CP and NCP steps can be combined to achieve more robust and performant samplers. Our contribution is to make the interleaving automatic and model-agnostic: instead of requiring the user to write multiple versions of their model and a custom inference algorithm, we implement iHMC as a black-box inference algorithm for centred Edward2 models.

\captionsetup{font=footnotesize,labelfont=footnotesize}
\begin{figure*}
\begin{multicols}{2}
\begin{algorithm}[H]        
	\captionof{algorithm}{ Interleaved Hamiltonian Monte Carlo }
	\label{alg:ihmc}     
	\fontsize{9pt}{1pt}\selectfont
	\begin{algorithmic}[1]
		\Arguments{ data $\mathbf{x}$; a centred model $M_{cp}(\mathbf{z} \mid \mathbf{x})$} 
		\Returns{ $S$ samples $\mathbf{z}^{(1)}, \dots \mathbf{z}^{(S)}$ from $p(\mathbf{z} \mid \mathbf{x})$}	
		\State $M_{ncp}(\mathbf{\tilde{z}} \mid \mathbf{x}), f = \kwsmall{make_ncp}(M_{cp}(\mathbf{z} \mid \mathbf{x}))$
		\State $\log p_{cp} = \kwsmall{make_log_joint}(M_{cp}(\mathbf{z} \mid \mathbf{x}))$
		\State $\log p_{ncp} = \kwsmall{make_log_joint}(M_{ncp}(\mathbf{\tilde{z}} \mid \mathbf{x}))$
		\State $ $ 
		\State $\mathbf{z}_0 = \kwsmall{init}() $ 
		\For{$s \in [1, \dots, S]$}
		\State $\quad\mathbf{z}' = \kwsmall{hmc_step}(\log p_{cp}, \mathbf{z}^{(s-1)})$ 
		\State $\quad\mathbf{z}'' = \kwsmall{hmc_step}(\log p_{ncp}, f^{-1}(\mathbf{z}'))$ 
		\State $\quad\mathbf{z}^{(s)} = f(\mathbf{z}'')$
		\EndFor
	    \State $ \vspace{5pt} $
		\Return $\mathbf{z}^{(1)}, \dots, \mathbf{z}^{(S)}$
	\end{algorithmic}
	\line
\end{algorithm}

\begin{algorithm}[H]
	\captionof{algorithm}{ Variationally Inferred Parameterisation }
	\label{alg:vip}     
	\fontsize{9pt}{1pt}\selectfont
	\begin{algorithmic}[1]
		\Arguments{ data $\mathbf{x}$; a centred model $M_{cp}(\mathbf{z} \mid \mathbf{x})$ } 
		\Returns{ $S$ samples $\mathbf{z}^{(1)}, \dots \mathbf{z}^{(S)}$ from $p(\mathbf{z} \mid \mathbf{x})$}
		\State $M_{vip}(\mathbf{\tilde{z}} \mid \mathbf{x}; \pparams), f = \kwsmall{make_vip}(M_{cp}(\mathbf{z} \mid \mathbf{x}))$
		\State $\log p(\mathbf{x}, \mathbf{\tilde{z}}) = \kwsmall{make_log_joint}(M_{vip} (\mathbf{\tilde{z}} \mid \mathbf{x}; \pparams))$
		\State $ $
		\State $Q(\mathbf{\tilde{z}}; \vparams) = \kwsmall{make_variational}(M_{vip}(\mathbf{\tilde{z}} \mid \mathbf{x}; \pparams))$
		\State $\log q(\mathbf{\tilde{z}}; \vparams) = \kwsmall{make_log_joint}(q(\mathbf{\tilde{z}}; \vparams))$ 
		\State $ $ 
		\State $\loss(\vparams, \pparams) = \mathbb{E}_{q}(\log p(\mathbf{x}, \mathbf{\tilde{z}}; \pparams)) - \mathbb{E}_{q}(\log q(\mathbf{\tilde{z}}; \vparams))$		
		\State $\vparams^*, \pparams^* = \displaystyle\argmax\loss(\vparams, \pparams)$
		\State $\log p(\mathbf{x}, \mathbf{\tilde{z}}) = \kwsmall{make_log_joint}(M_{vip} (\mathbf{\tilde{z}} \mid \mathbf{x}; \pparams^*))$
		
		\State $ \mathbf{z}^{(1)}, \dots, \mathbf{z}^{(S)} = \kwsmall{hmc}(\log p)$\\
		\Return $f(\mathbf{z}^{(1)}), \dots, f(\mathbf{z}^{(S)})$
	\end{algorithmic}
	\line
\end{algorithm}
\end{multicols}\vspace{-18pt}
\end{figure*}
\captionsetup{font=small,labelfont=small}

Algorithm~\ref{alg:ihmc} outlines iHMC. It takes a single centred model $M_{cp}(\mathbf{z} \mid \mathbf{x})$ that defines latent variables $\mathbf{z}$ and generates data $\mathbf{x}$. It uses the function \lstinline{make_ncp} to automatically obtain a non-centred version of the model, $M_{ncp}(\mathbf{\tilde{z}} \mid \mathbf{x})$, which defines auxiliary variables $\mathbf{\tilde{z}}$ and function $f$, such that $\mathbf{z} = f(\mathbf{\tilde{z}})$. 

\subsection{Variationally inferred parameterisation} \label{ssec:vip}

\begin{figure}
\centering
\begin{subfigure}[b]{\linewidth}
    \centering
	\includegraphics[width=0.9\linewidth]{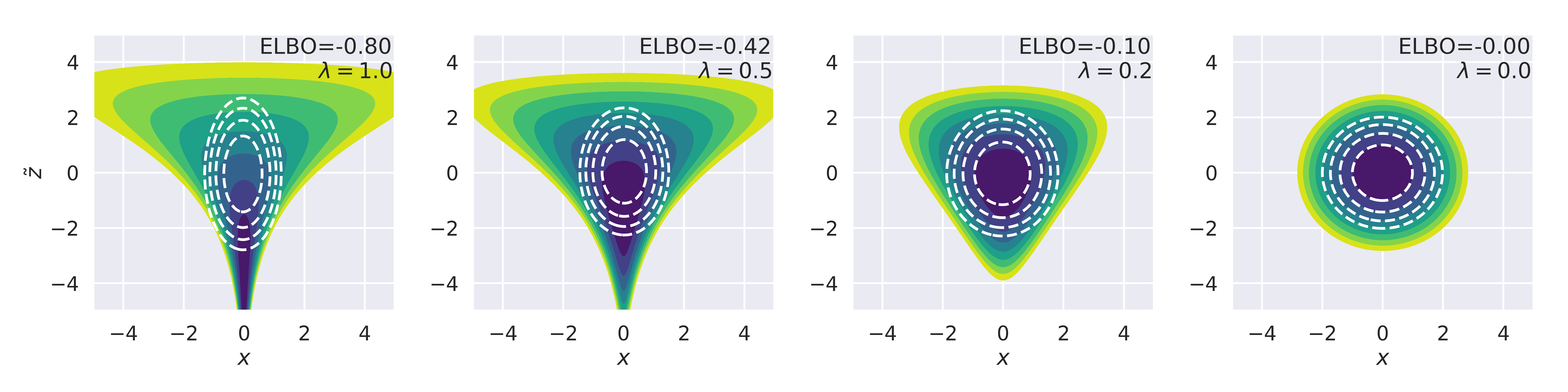}
    \vspace{-6pt}
	\caption{Different parameterisations $\pparams$ of the funnel, with mean-field normal variational fit $q(\tilde{\mathbf{z}})$(overlayed in white).}
	\label{fig:funnel_transformation}
\end{subfigure}
\vspace{-2pt}

\begin{subfigure}[b]{\linewidth}
    \centering
	\includegraphics[width=0.9\linewidth]{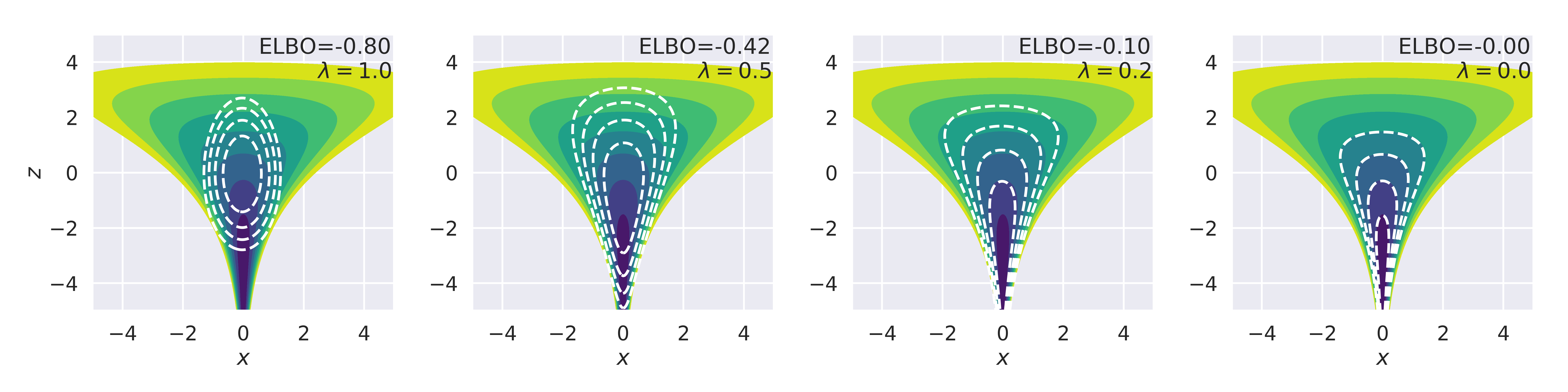}
	\vspace{-6pt}
	\caption{Alternative view as implicit variational distributions $q_{\pparams}^*(\mathbf{z})$ (overlayed in white) on the original space.}
	\label{fig:funnel_transformation_q}
\end{subfigure}

\caption{Neal's funnel: $z \sim N(0, 3);$ $x \sim N(0, e^{z/2})$, with mean-field normal variational fit overlayed.}
\vspace{-12pt}
\end{figure}

The best parameterisation for a given model may mix centred and non-centred representations for different variables. 
To efficiently search the space of reparameterisations, we propose the \textit{variationally inferred parameterisation (VIP)} algorithm,  which selects a parameterisation by gradient-based optimisation of a differentiable variational objective. VIP can be used as a pre-processing step to another inference algorithm; as it only changes the parameterisation of the model, MCMC methods applied to the learned parameterisation maintain their asymptotic guarantees.

Consider a model with latent variables $\mathbf{z}$. 
We introduce \emph{parameterisation parameters} 
$\pparams=(\lambda_i) \in [0, 1]$ for each variable $z_i$, and transform $z_i \sim \normal(z_i \mid \mu_i, \sigma_i)$ by defining $\tilde{z}_i \sim \normal(\lambda_i\mu_i, \sigma_i^{\lambda_i})$ and $z_i = \mu_i + \sigma_i^{1 - \lambda_i}(\tilde{z}_i - \lambda_i\mu_i)$.
This defines a continuous relaxation that includes NCP as the special case $\pparams=0$ and CP as $\pparams=1$. More generally, it supports a combinatorially large class of per-variable and partial centrings. We aim to choose the parameterisation $\pparams$ under which the posterior $p(\tilde{\mathbf{z}} | \mathbf{x}; \pparams)$ is ``most like'' an independent normal distribution.

\begin{figure}
	\vspace{4pt}
	\flushleft \hspace{8pt}
	\includegraphics[width=0.8\linewidth]{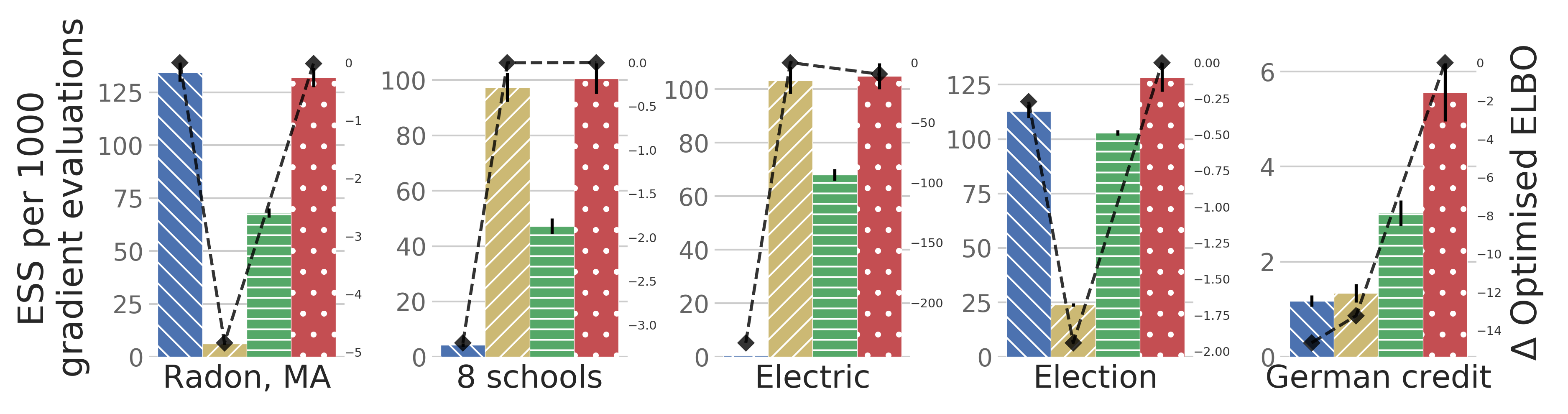}
	\caption{Effective sample size (w/ 95\% intervals)  and the optimised ELBO across several models.}
	\label{fig:ess_and_elbo_results}
	\vspace{-12pt}
\end{figure}

A natural objective to minimise is $\textrm{KL}(q(\tilde{\mathbf{z}}; \vparams) \mid\mid p(\tilde{\mathbf{z}} \mid \mathbf{x}; \pparams))$, where $q(\mathbf{\tilde{z}}; \vparams) = \normal(\tilde{\mathbf{{z}}} \mid \boldsymbol{\mu}, \textrm{diag}(\boldsymbol{\sigma}))$ is an independent normal model with \textit{variational parameters} $\vparams = (\boldsymbol{\mu}, \boldsymbol{\sigma})$. Minimising this divergence corresponds to maximising a variational lower bound, the ELBO \citep{bishop2006pattern}:
\[\loss(\vparams, \pparams) = \mathbb{E}_{q(\tilde{\mathbf{z}}; \vparams)}\left(\log p(\mathbf{x}, \mathbf{\tilde{z}}; \pparams) - \log q(\mathbf{\tilde{z}}; \vparams)\right) \le \log p(\mathbf{x}).\]
Note that the auxiliary parameters $\pparams$ are not statistically identifiable: the marginal likelihood $\log p(\data; \pparams) = \log p(\data)$ is constant with respect to $\pparams$.  
However, the computational properties of the reparameterised models differ, which is what the variational bound selects for. Our key hypothesis (which the results in Figure~\ref{fig:ess_and_elbo_results} seem to support) is that diagonal-normal approximability is a good proxy for MCMC sampling efficiency.

To search for a good model reparameterisation, we optimise $\mathcal{L}(\vparams, \pparams)$ using stochastic gradients to simultaneously fit the variational distribution $q$ to the posterior $p$ and optimise the shape of that posterior.
\autoref{fig:funnel_transformation} provides a visual example: an independent normal variational distribution is a poor fit to the pathological geometry of a centred Neal's funnel, but non-centring leads to a well-conditioned posterior, where the variational distribution is a perfect fit. 
In general settings where the reparameterised model is not exactly Gaussian, sampling-based inference can be used to refine the posterior; we apply VIP as a preprocessing step for HMC (summarised in Algorithm~\ref{alg:vip}). Both the reparameterisation and the construction of the variational model $q$ are implemented as automatic program transformations using Edward2's interceptors.

An alternate interpretation of VIP is that it expands a variational family to a more expressive family capable of representing prior dependence. Letting $\tilde{\mathbf{z}} = f_{\pparams}(\mathbf{z})$ represent the partial centring transformation, an independent normal family $q(\tilde{\mathbf{z}})$ on the transformed model corresponds to an implicit posterior $q_{\pparams}^*(\mathbf{z}) = q\left(\tilde{\mathbf{z}} = f_{\pparams}(\mathbf{z})\right)|f'_{\pparams}(\mathbf{z})|^{-1}$ on the original model variables. Under this interpretation, $\pparams$ are variational parameters that serve to add freedom to the variational family, allowing it to interpolate from independent normal (at $\lambda_i=1$, \autoref{fig:funnel_transformation_q} left) to a representation that captures the exact prior dependence structure of the model (at $\lambda_i=0$, \autoref{fig:funnel_transformation_q} right).

\section{Experiments}

\begin{figure}
	\centering
	\begin{subfigure}{0.8\linewidth}
		\includegraphics[width=\linewidth]{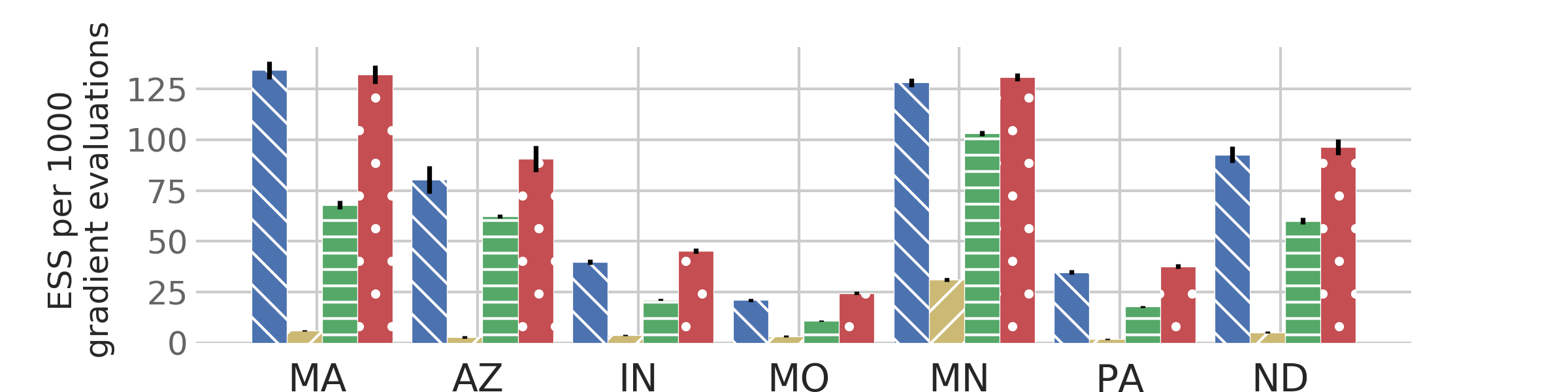}
	\end{subfigure}
	\begin{subfigure}[c]{0.18\linewidth}
		\includegraphics[width=\linewidth]{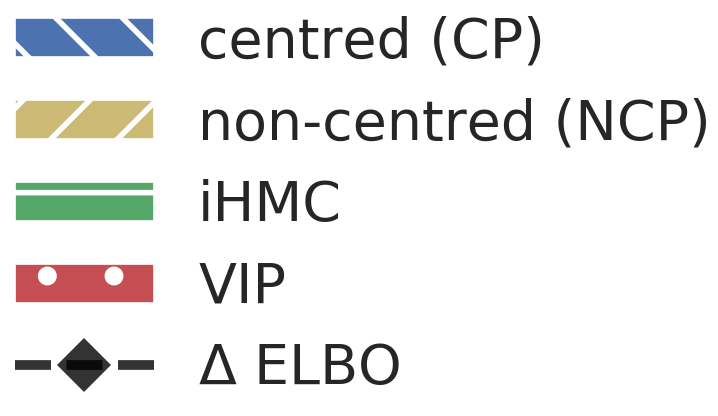}
	\end{subfigure}
	\caption{Effective sample size and 95\% confidence intervals for the radon model across US states.}
    \label{fig:radon_results}
    \vspace{-12pt}
\end{figure}

We evaluate our proposed approaches by using Hamiltonian Monte Carlo to sample from the posterior of hierarchical Bayesian models on several datasets:

\textbf{Eight schools} \cite{8Schools}: estimating the treatment effects $\theta_i$ of a course taught at each of $i = 1\ldots 8$ schools, given test scores $y_i$ and standard errors $\sigma_i$:
\[
\mu \sim \normal(0, 5)\qquad
\log \tau \sim \normal(0, 5)\qquad
\theta_i \sim \normal(\mu, \tau)\qquad
y_i \sim \normal(\theta_i, \sigma_i)
\]
\textbf{Radon} \cite{GelmanAndHill}: hierarchical linear regression, in which the radon level $r_i$ in a home $i$ in county $c$ is modelled as a function of the  (unobserved) county-level effect $m_c$, the county uranium reading $u_c$, and $x_i$, the number of floors in the home:
\[
\mu, a, b \sim \normal(0, 1)\qquad
m_c \sim \normal(\mu + a u_c, 1)\qquad
\log r_i \sim \normal(m_{c[i]} + b x_i, \sigma)
\]
\textbf{German credit} \citep{Dua2019}: logistic regression; hierarchical prior on coefficient scales:
\[
\log \tau_0 \sim \normal(0, 10)\qquad
\log \tau_i \sim \normal(\log \tau_0, 1)\qquad
\beta_i \sim \normal(0, \tau_i)\qquad
y \sim \text{Bernoulli}(\sigma(\beta X^T))
\]
\textbf{Election '88} \cite{GelmanAndHill}: logistic model of 1988 US presidential election outcomes by county, given demographic covariates $\mathbf{x}_i$ and state-level effects $\alpha_s$:
\[
\beta_d \!\sim\! \normal(0, 100)\;\;\,
\mu \!\sim\! \normal(0, 100)\;\;\,
\log \tau \!\sim\! \normal(0, 10)\;\;\,
\alpha_s \!\sim\! \normal(\mu, \tau)\;\;\,
y_i \!\sim\! \text{Bernoulli}(\sigma(\alpha_{s[i]} + \beta^T \mathbf{x}_i))
\]
\textbf{Electric Company} \citep{GelmanAndHill}: paired causal analysis of the effect of viewing an educational TV show on each of $192$ classforms over $G = 4$ grades. The classrooms were divided into $P=96$ pairs, and one class in each pair was treated ($x_i=1$) at random:
\[
\mu_g \!\sim\! \normal(0, 1)\;\;\;
a_p \!\sim\! \normal (\mu_{g[p]}, 1)\;\;\;
b_g \!\sim\! \normal(0, 100)\;\;\;
\log \sigma_g \!\sim\! \normal(0, 1)\;\;\;
y_i \!\sim\! \normal(a_{p[i]} + b_{g[i]} x_i, \sigma_{g[i]} )
\]

\subsection{Algorithms and experimental details}

For each model and dataset, we compare our methods, interleaved HMC (iHMC) and VIP-HMC, with baselines of running HMC on either fully centred (CP-HMC) or fully non-centred (NCP-HMC) models. We initialise each HMC chain with samples from an independent Gaussian variational posterior, and use the posterior scales as a diagonal preconditioner; for VIP-HMC this variational optimisation also includes the parameterisation parameters $\pparams$. All variational optimisations were run for the same number of steps, so they were a fixed cost across all methods except iHMC (which depends on preconditioners for both the centred and non-centred transition kernels). The HMC step size and number of leapfrog steps were tuned following the procedures described in \autoref{ap:experiments}.

We report the average effective sample size per $1000$ gradient evaluations ($\ess$), with standard errors computed from $200$ chains. We use gradient evaluations, rather than wallclock time, as they are the dominant operation in both HMC and VI and are easier to measure reliably; in practice, the wallclock times we observed per gradient evaluation did not differ significantly between methods. 

Full details on the set up of the experiments can be found in \autoref{ap:experiments}.

\subsection{Results}

Figures \ref{fig:radon_results} and \ref{fig:ess_and_elbo_results} show the results of the experiments.
In most cases, either the centred or non-centred parameterisation works well, while the other does not. An exception is the German credit dataset, where both CP-HMC and NCP-HMC give a small ESS: $1.2 \pm 0.2$ or $1.3 \pm 0.2~\ess$ respectively. 

\paragraph{iHMC.} 
Across the datasets in both figures, we see that iHMC is a robust alternative to CP-HMC and NCP-HMC. Its performance is always within a factor of two of the best of CP-HMC and NCP-HMC, and sometimes better.
In addition to being robust, iHMC can sometimes navigate the posterior more efficiently than either of CP-HMC and NCP-HMC can: in the case of German credit, it performs better than both ($3.0 \pm 0.2~\ess$).

\begin{figure}[t]
	\centering
	\includegraphics[width=\linewidth]{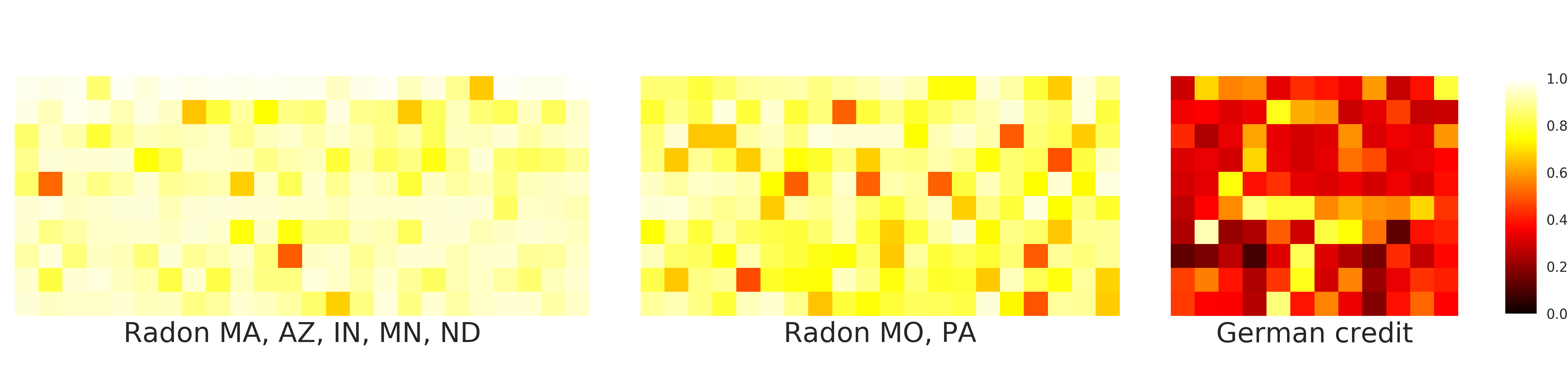}
	\caption{A heat map of VIP parameterisations. Light regions correspond to CP and dark regions to NCP.}
	\label{fig:parametrisation_results}
	\vspace{-12pt}
\end{figure}

\paragraph{VIP.}
Performance of VIP-HMC is typically as good as the better of CP-HMC and NCP-HMC, and sometimes better. On the German credit dataset, it achieves $5.6 \pm 0.6~\ess$, more than three times the rate of CP-HMC and NCP-HMC, and significantly better than iHMC.
\autoref{fig:ess_and_elbo_results} shows the correspondence between the optimised mean-field ELBO and the effective sampling rate. This result supports the ELBO as a reasonable predictor of the conditioning of a model, and further confirms the validity of VIP as a strategy for automatic reparameterisation. 
Finally, we show some of the parameterisations that VIP finds in \autoref{fig:parametrisation_results}. VIP's behaviour appears reasonable: for most datasets we looked at, VIP finds the ``correct'' global parameterisation: most parameterisation parameters are set to either $0$ or $1$ (\autoref{fig:parametrisation_results}, left). In the cases where a global parameterisation is not optimal (e.g. radon MO, radon PA and, most notably, German credit), VIP finds a mixed parameterisation, combining centred, non-centred, and partially centred variables (\autoref{fig:parametrisation_results}, centre and right). 

\section{Discussion}

Our results demonstrate that automated reparameterisation of probabilistic models is practical, and enables inference algorithms that can in some cases find parameterisations even better than those a human could realistically express. These techniques allow modellers to focus on expressing statistical assumptions, leaving computation to the computer. We view the methods in this paper as exciting proofs of concept, and hope that they will inspire additional work in this space. 

While we focus on reparameterising hierarchical models naturally written in centred form, the inverse transformation---detecting and exploiting implicit hierarchical structure in models expressed as algebraic equations---is an important area of future work. This may be compatible with recent trends exploring the use of symbolic algebra systems in PPL runtimes \cite{narayanan2016probabilistic, autoconj2018}. We also see promise in automating reparameterisations of heavy-tailed and multivariate distributions, and in designing new inference algorithms to exploit these capabilities.

\newpage
\bibliographystyle{apalike}
\bibliography{bib}

\cleardoublepage
\appendix

\section{Derivation of the condition number of the posterior for a simple model} \label{ap:maths}

\paragraph{Centred parameterisation} 
\[\theta \sim \normal(0, 1) \qquad \mu \sim \normal(\theta, \sigma_{\mu}) \qquad y_n \sim \normal(\mu, \sigma) \text{ for all } n \in 1 \dots N\]

\paragraph{Non-centred parameterisation} 
\[\theta \sim \normal(0, 1) \qquad \tilde{\mu} \sim \normal(0, 1) \qquad y_n \sim \normal(\theta + \sigma_{\mu}\tilde{\mu}, \sigma) \text{ for all } n \in 1 \dots N\]

As the Gaussian distribution is self-conjugate, the posterior distribution (given $\data$) in each case (centred or non-centred) is also a Gaussian distribution, whose shape is entirely specified by a covariance matrix $V$.
To quantify the quality of each parameterisation, we investigate the condition number $\kappa$ of the posterior covariance matrix in each case under the best diagonal preconditioner. 

We do this in three steps:  
\begin{enumerate}
\item We derive the covariance matrices $V_{\alg{cp}}$ and $V_{\alg{ncp}}$, such that $p(\mu, \theta \mid \mathbf{y}) = \normal\left(\mu, \theta \mid \mathbf{m}_{\alg{cp}}, V_{\alg{cp}}\right)$ and $p(\tilde{\mu}, \theta \mid \mathbf{y}) = \normal\left(\tilde{\mu}, \theta \mid \mathbf{m}_{\alg{ncp}}, V_{\alg{ncp}}\right)$ (\autoref{eq:v_cp} and \autoref{eq:v_ncp}).

\item We find the best diagonal preconditioners $D^*_{\alg{cp}}$ and $D^*_{\alg{ncp}}$: for $\alg{p} = \alg{cp}, \alg{ncp}$, that is $D_\alg{p}^* = \arg\min_D ( \lambda_\alg{p}^{(2)} / \lambda_\alg{p}^{(1)} )$,  where $\lambda_\alg{p}^{(1)}$ and $\lambda_\alg{p}^{(2)}$ are the eigenvalues of $U = D^T V_\alg{p} D$ (\autoref{eq:d_cp} and \autoref{eq:d_ncp}). 

\item We compare the condition numbers $\kappa_{cp}(q) = \lambda_{cp}^{(2)} / \lambda_{cp}^{(1)}$ and $\kappa_{ncp}(q) = \lambda_{ncp}^{(2)} / \lambda_{ncp}^{(1)}$, where $\lambda_{(n)cp}^{(i)}$ are the eigenvalues of $U^* = (D^*)^T V D^*$
\end{enumerate}

\subsection{Deriving $V_{\alg{cp}}$ and $V_{\alg{ncp}}$: centred parameterisation}

\begin{align*}
p(\mu, \theta \mid \mathbf{y}) \propto\;& p(\mu, \theta, \mathbf{y}) \\
\propto\;& \mathcal{N}(\mu \mid \theta, \sigma_{\mu})\mathcal{N}(\theta \mid 0, 1)\prod_{n=1}^N\mathcal{N}( y_n \mid \mu, \sigma) \\
\propto\;& \exp\left(-\frac{1}{2}\left( \frac{(\mu - \theta)^2}{\sigma_{\mu}^2} + \theta^2 + \sum_{n=1}^N\frac{(y_n - \mu)^2}{\sigma^2}  \right)\right) \\
\propto\;& \exp\left(-\frac{1}{2}\left(   \mu^2\left(\frac{1}{\sigma_{\mu}^2} + \frac{N}{\sigma^2}\right) + \theta^2 \left(\frac{1}{\sigma_{\mu}^2} + 1\right) - 2\mu\theta \left( \frac{1}{\sigma_{\mu}^2}\right) + \mu\left(\frac{-2}{\sigma^2}\sum_{n=1}^Ny_n\right)  \right)\right)
\end{align*}

At the same time, for $A = V_{\alg{ncp}}^{-1}$, we have:

\begin{align*}
\mathcal{N}(\mu, \theta \mid \mathbf{m}_{\alg{cp}}, V_{\alg{cp}}) \propto{}&   \exp\left(- \frac{1}{2}  \left(\binom{\mu}{\theta} - \mathbf{m}\right)^T A \left(\binom{\mu}{\theta} - \mathbf{m}\right) \right) \\
\begin{split}
\propto{}&  \exp\left( -\frac{1}{2} \left(  \mu^2A_{11} + \theta^2A_{22} + \mu\theta(2A_{12}) + \mu(-2A_{11}m_1 - 2A_{12}m_2) + \right.\right.\\
&{}\qquad \left.\vphantom{\frac{1}{2}}\left. \mu^2A_{11}  \theta(-2A_{22}m_2 - 2A_{12}m_1)   \right)\right)
\end{split}
\end{align*}

Thus, for $q = N / \sigma^2$, we get: 
$A = \left({\begin{array}{cc} \frac{1}{\sigma_{\mu}^2} + q  & - \frac{1}{\sigma_{\mu}^2} \\ - \frac{1}{\sigma_{\mu}^2} & \frac{1}{\sigma_{\mu}^2} + 1 \end{array}}\right)$
And therefore:
\begin{equation} \label{eq:v_cp}
V_{\alg{cp}} = \frac{1}{\sigma_{\mu}^2q + q+1}\left( {\begin{array}{cc} 1 + \sigma_{\mu}^2 & 1 \\  1 & q\sigma_{\mu}^2 + 1 \\ \end{array} } \right)
\end{equation}

\subsection{Deriving $V_{\alg{cp}}$ and $V_{\alg{ncp}}$: non-centred parameterisation}
Like in the previous subsection, we have:

\begin{align*}
p(\epsilon, \theta \mid \mathbf{y}) \propto\;& p(\epsilon, \theta, \mathbf{y}) \\
\propto\;& \mathcal{N}(\epsilon \mid 0, 1)\mathcal{N}(\theta \mid 0, 1)\prod_{n=1}^N\mathcal{N}( y_n \mid \sigma_{\mu}\epsilon+\theta, \sigma) \\
\propto\;& \exp\left(-\frac{1}{2}\left(  (\epsilon^2 + \theta^2 + \sum_{n=1}^N\frac{(y_n - \sigma_{\mu}\epsilon - \theta)^2}{\sigma^2}  \right)\right) \\
\propto\;& \exp\left(-\frac{1}{2}\left(   \epsilon^2\left(1 + \frac{N\sigma_{\mu}^2}{\sigma^2}\right) + \theta^2\left(1 + \frac{N}{\sigma^2}\right) + \epsilon\theta\left(  \frac{2N\sigma_{\mu}}{\sigma^2} \right) + \right.\right.\\
&{}\qquad\qquad \left.\left.  \vphantom{\frac{1}{2}} \epsilon\left(  \frac{-2\sigma_{\mu}\sum y_n}{\sigma^2} \right) + \theta\left(  \frac{-2 \sum y_n}{\sigma^2} \right)  \right)\right)
\end{align*} 

Similarly to before, we derive $A = \left({\begin{array}{cc} \sigma_{\mu}^2q + 1 & \sigma_{\mu}q \\ \sigma_{\mu}q & q+1 \end{array}}\right)$, and therefore:
\begin{equation} \label{eq:v_ncp}
V_{\alg{ncp}} = \frac{1}{\sigma_{\mu}^2q + q+1}\left( {\begin{array}{cc} q+1& -\sigma_{\mu}q \\ -\sigma_{\mu}q & \sigma_{\mu}^2q+1 \\ \end{array} } \right)
\end{equation}

\subsection{The best diagonal preconditioner}

Consider a diagonal preconditioner $D = \left({\begin{array}{cc} d & 0 \\ 0 & 1 \end{array}} \right)$. The best diagonal preconditioner $D^*$ of $V$ is such that:
\[D^* = \underset{D}{\arg\min} \left( \lambda_2 / \lambda_1 \right) \text{ where } \lambda_1, \lambda_2 \text{ are the eigenvalues of } U = D^T V D \]

Firstly, in terms of the covariance matrix in the centred case, we have:
\begin{align*}
U = D^T V_{\alg{cp}} D &= 
\left({\begin{array}{cc} d & 0 \\ 0 & 1 \end{array}} \right) 
\left(\frac{1}{\sigma_{\mu}^2q + q+1}\left( {\begin{array}{cc} 1 + \sigma_{\mu}^2 & 1 \\  1 & q\sigma_{\mu}^2 + 1 \\ \end{array} } \right)\right) 
\left({\begin{array}{cc} d & 0 \\ 0 & 1 \end{array}} \right) \\
&= \frac{1}{\sigma_{\mu}^2q + q+1}\left( {\begin{array}{cc} (1 + \sigma_{\mu}^2)d^2 & d \\  d & q\sigma_{\mu}^2 + 1 \\ \end{array} } \right)
\end{align*}

\newcommand{\p}{(\sigma_{\mu}^2q + q+1)}
\newcommand{\s}{\sigma_{\mu}^2}

The solutions of $det(U-\lambda I) = 0$ are the solutions of:
\[
((1 + \sigma_{\mu}^2)d^2 - \lambda(\sigma_{\mu}^2q + q+1))(q\sigma_{\mu}^2 + 1 - \lambda(\sigma_{\mu}^2q + q+1)) - d^2 = 0
\]
which, after simplification, becomes:
\[
\p\lambda^2 - (\sigma_\mu^2q + 1 + d^2(\sigma_\mu^2 + 1))\lambda + d^2\sigma_\mu^2 = 0
\]

We want to find $d$ that minimises $\lambda_2 / \lambda_1$. Let $u = d^2$. We are looking for $u$, such that
$\frac{\partial}{\partial u} \frac{\lambda_2}{\lambda_1} = 0$, in order to find $d^*_{\alg{cp}} = \underset{d}{\arg\min} \left( \lambda_2 / \lambda_1 \right)$.
By expanding and simplifying we get:
\[
2 \frac{\partial}{\partial u} (\sigma_\mu^2q + 1 + u(\sigma_\mu^2 + 1)) = (\sigma_\mu^2q + 1 + u(\sigma_\mu^2 + 1)) / u
\]

And thus:
\begin{equation} \label{eq:d_cp}
	d_{\alg{cp}}^* = \sqrt{u} = \sqrt{\frac{\s q + 1}{\s + 1}}
\end{equation}

We obtain the best diagonal preconditioner $D^*_{\alg{ncp}} = \left({\begin{array}{cc} d^*_{\alg{ncp}} & 0 \\ 0 & 1 \end{array}} \right)$ in a similar manner, finally getting:
\begin{equation} \label{eq:d_ncp}
d_{\alg{ncp}}^* = \sqrt{u} = \sqrt{\frac{\s q + 1}{q + 1}}
\end{equation}

\subsection{The condition numbers $\kappa_{\alg{cp}}$ and $\kappa_{\alg{ncp}}$}

Finally, we substitute $d^*_{\alg{cp}}$ and $d^*_{\alg{ncp}}$ in the respective eigenvalue equations to derive the condition number in each case:
\begin{equation}
\kappa_{\alg{cp}} = \lambda_2^{(\alg{cp})} / \lambda_1^{(\alg{cp})} =
\frac{\s q + 1 + \sqrt{(\s q + 1)^2 - \s (\s q + q + 1)  (\s q + 1) / (v + 1)}}{\s q + 1 - \sqrt{(\s q + 1)^2  - \s (\s q + q + 1)  (\s q + 1) / (v + 1)}}
\end{equation}

\begin{equation}
\kappa_{\alg{ncp}} = \lambda_2^{(\alg{ncp})} / \lambda_1^{(\alg{ncp})} =
\frac{\s q + 1 + \sqrt{(\s q + 1)^2 - \s (\s q + q + 1)  (\s q + 1) / (q + 1)}}{\s q + 1 - \sqrt{(\s q + 1)^2  - \s (\s q + q + 1)  (\s q + 1) / (q + 1)}}
\end{equation}

\section{Interceptors} \label{ap:interceptors}

Interceptors can be used as a powerful abstractions in a probabilistic programming systems, as discussed previously by \citet{ProbProg18}, and shown by both Pyro and Edward2. In particular, we can use interceptors to automatically reparameterise a model, as well as to specify variational families. In this section, we show Edward2 pseudo-code for the interceptors used to implement iHMC and VIP-HMC.

\subsection{Make log joint} \label{ap:make_log_joint}

The following code is an outline of Edward2's impllementation of a function that evaluates the log density $\log p(\mathbf{x})$ at some given $\mathbf{x}$:
\begin{lstlisting}
def make_log_joint_fn(model):
	def log_joint_fn(**kwargs):
  		log_prob = 0

  		def log_prob_interceptor(rv_constructor, **rv_kwargs):
    		# Overrides a random variable's `value` and accumulates its log prob.
    		rv_name = rv_kwargs.get("name")
    		rv_kwargs["value"] = kwargs.get(rv_name)
      
    		rv = rv_constructor(**rv_kwargs)
    		log_prob = log_prob + rv.distribution.log_prob(rv.value)
    		return rv
        
  		with ed.interception(log_prob_interceptor):
    		model()
    
  		return log_prob
	return log_joint_fn
\end{lstlisting}

By executing the \lstinline{model} function in the context of \lstinline{log_prob_interceptor}, we override each sample statement (a call to a random variable constructor \lstinline{rv_constructor}), to generate a variable that takes on the value provided in the arguments of \lstinline{log_joint_fn}. As a side effect, we also accumulate the result of evaluating each variable's prior density at the provided value, which, by the chain rule, gives us the log joint density. 

\subsection{Non-centred Parameterisation Interceptor}
By intercepting every construction of a normal variable (or, more generally, of location-scale family variables), we can create a standard normal variable instead, and scale and shift appropriately.

\begin{lstlisting}
def ncp_interceptor(rv_constructor, **rv_kwargs):
	# Assumes rv_constructor is in the location-scale family
	name = rv_kwargs["name"] + "_std"
	rv_std = ed.interceptable$\footnotemark$(rv_constructor)(loc=0, scale=1)
	return rv_kwargs["loc"] + rv_kwargs["scale"] * rv_std
\end{lstlisting}
\footnotetext{Wrapping the constructor in with \lstinline{ed.interceptable} ensures that we can nest this interceptor in the context of other interceptors.}

Running a model that declares the random variables $\boldsymbol{\theta}$ in the context of \lstinline{ncp_interceptor} will declare a new set of standard normal random variables $\boldsymbol{\theta}^{(\mathrm{std})}$. Nesting this in the context of the \lstinline{log_prob_interceptor} from \autoref{ap:edward} will then evaluate the log joint density $\log p(\boldsymbol{\theta}^{(\mathrm{std})})$. 

For example, going back to Neal's funnel, running
\begin{lstlisting}
with ed.interception(log_prob_interceptor):
	neals_funnel()
\end{lstlisting}
corresponds to evaluating $\log p(z, x) = \log \normal(z \mid 0, 3) + \log \normal(x \mid 0, e^{z / 2})$, while running 
\begin{lstlisting}
with ed.interception(log_prob_interceptor):
	with ed.interception(ncp_interceptor):
		neals_funnel()
\end{lstlisting}
corresponds to evaluating $\log p(z^{(\mathrm{std})}, x^{(\mathrm{std})}) = \log \normal(z^{(\mathrm{std})} \mid 0, 1) + \log \normal(x^{(\mathrm{std})} \mid 0, 1)$.

\subsection{VIP Interceptor}
The VIP interceptor is similar to the NCP interceptor. The notable difference is that it creates new learnable Tensorflow variables, which correspond to the parameterisation parameters $\pparams$: 
\begin{lstlisting}
def vip_interceptor(rv_constructor, **rv_kwargs):
	name = rv_kwargs["name"] + "_vip"
	rv_loc = rv_kwargs["loc"]
	rv_scale = rv_kwargs["scale"]
	
	a = tf.nn.sigmoid(tf.get_variable(
	  name + "_a_unconstrained",
	  initializer=tf.zeros_like(rv_loc))
	
	rv_vip = ed.interceptable(rv_constructor)(
	           loc=a * rv_loc, scale=rv_scale ** a)
	return rv_loc + rv_scale ** (1 - a) * (rv_vip - a * rv_loc)

\end{lstlisting}

\subsection{Mean-field Variational Model Interceptor}
Finally, we show a mean-field variational familiy interceptor, which we use both to tune the step sizes for HMC (see \autoref{ap:experiments}), and to make use of VIP automatically. The \lstinline{mfvi_interceptor} simply substitutes each sample statement with sampling from a normal distribution with parameters specified by some fresh variational parameters $\mu$ and $\sigma$:
\begin{lstlisting}
def vip_interceptor(rv_constructor, **rv_kwargs):
	name = rv_kwargs["name"] + "_q"
	mu = tf.get_variable(name + "_mu")
	sigma = tf.nn.softmax(tf.get_variable(name + "_sigma"))
	
	rv_q = ed.interceptable(ed.Normal)(
	         loc=mu, scale=sigma, name=name)
	return rv_q

\end{lstlisting}

\section{Details of the experiments} \label{ap:experiments}

\paragraph{Algorithms.}
\begin{itemize}
	\item CP-HMC: HMC run on a fully centred model.
	\item NCP-HMC: HMC run on a fully non-centred model.
	\item iHMC: interleaved HMC.
	\item VIP-HMC: HMC run on the a model reparameterised as given by VIP.
\end{itemize}

Each run consists of VI pre-processing and HMC inference.

\paragraph{Variational inference pre-processing.}

We use automatic differentiation to compute stochastic gradients of the ELBO with respect to $\pparams, \vparams$ and perform the optimisation using Adam \cite{kingma2014adam}. We implement the constraint $\lambda_i \in [0, 1]$ using a sigmoid transformation; $\lambda_i = 1/\left(1 + \exp(-\tilde{\lambda_i})\right)$ for $\tilde{\lambda_i} \in \mathbb{R}$.

Prior to running HMC, we also run VI to approximate per-variable initial step sizes (equivalently, a diagonal preconditioning matrix), and to initialise the chains. For each of CP-HMC and NCP-HMC this is just mean-field VI, and for VIP-HMC the VI procedure is VIP.

Each VI method is run for $3000$ optimisation steps, and the ELBO is approximated using $256$ Monte Carlo samples. We use the Adam optimiser with initial learning rate $\alpha \in [0.02, 0.05, 0.1, 0.2, 0.4]$, decayed to $\alpha / 5$ after 1000 steps and $\alpha / 20$ after 2000 steps, and returned the result with the highest ELBO. 

\paragraph{Hamiltonian Monte Carlo inference.}
In each case we run $200$ chains for a warm-up period of $2000$ steps, followed by $10000$ steps each, and 
report the average effective sample size (ESS) per $1000$ gradient evaluations ($\ess$). Since ESS is naturally estimated from scalar traces, we first estimate per-variable effective sample sizes for each model variable, and take the overall ESS to be the minimum across all variables. 

The HMC step size $s_t$ was adapted to target an acceptance probability of 0.75, following a simple update rule
\[\log s_{t_+1} = \log s_t + 0.02 \cdot (\mathbb{I}[\alpha_t - 0.75] - \mathbb{I}[0.75 - \alpha_t])\]
where $\alpha_t$ is the acceptance probability of the proposed state at step $t$ \citep{andrieu2008tutorial}. The adaptation runs during the first 1500 steps of the warm-up period, after which we allow the chain to mix towards a stationary distribution.

The number of leapfrog steps is chosen using `oracle' tuning: each sampler is run with logarithmically increasing number of leapfrog steps in $\{1, 2, 4, \ldots, 128\}$, and we report the result that maximises $\ess$. This is intended to decouple the problem of tuning the number of leapfrog steps from the issues of parameterisation consider in this paper, and ensure that each method is reasonably tuned. For iHMC, we tune a single number of leapfrog steps that is shared across both the CP and NCP substeps.

\end{document}